# Arabic Language Sentiment Analysis on Health Services

Abdulaziz M. Alayba[1], Vasile Palade[2], Matthew England[3] and Rahat Iqbal[4]

*Faculty of Engineering, Environment and Computing*
*Coventry University*
*Coventry, UK*

[1] Alaybaa@uni.coventry.ac.uk
[2] Vasile.Palade@coventry.ac.uk
[3] Matthew.England@coventry.ac.uk
[4] R.Iqbal@coventry.ac.uk

*Abstract* — **The social media network phenomenon leads to a massive amount of valuable data that is available online and easy to access. Many users share images, videos, comments, reviews, news and opinions on different social networks sites, with Twitter being one of the most popular ones. Data collected from Twitter is highly unstructured, and extracting useful information from tweets is a challenging task. Twitter has a huge number of Arabic users who mostly post and write their tweets using the Arabic language. While there has been a lot of research on sentiment analysis in English, the amount of researches and datasets in Arabic language is limited.**

**This paper introduces an Arabic language dataset which is about opinions on health services and has been collected from Twitter. The paper will first detail the process of collecting the data from Twitter and also the process of filtering, pre-processing and annotating the Arabic text in order to build a big sentiment analysis dataset in Arabic. Several Machine Learning algorithms (Naïve Bayes, Support Vector Machine and Logistic Regression) alongside Deep and Convolutional Neural Networks were utilized in our experiments of sentiment analysis on our health dataset.**

## I. INTRODUCTION

In the past ten years, many social network sites (Facebook, Twitter, Instagram etc.) have increased the presence on the web. These sites have an enormous number of users who produce a massive amount of data which include texts, images, videos, etc. According to [1], the estimated amount of data on the web will be about 40 thousand Exabytes, or 40 trillion gigabytes, in 2020. Analysis of such date could be valuable. There are many different techniques for data analytics on data collected from the web, with sentiment analysis a prominent one. Sentiment analysis (see for example [2]) is the study of people's attitudes, emotions and options, and involves a combination of text mining and natural language processing. Sentiment analysis focuses on analyzing text messages that hold people opinions. Examples of topics for analysis include opinions on products, services, food, educations, etc. [3].

Twitter is a popular social media platform where a huge number of tweets are shared and many tweets contain valuable data. As [4] reported: in March 2014, active Arabic users wrote over 17 million tweets per day. There are huge numbers of tweets generated every minute and many of them in the Arabic language. Topics about health services appear frequently on Twitter trends. The aims of this paper is to introduce a new Arabic data set on health services for opinion mining purposes. Also, to explain the process of collecting data from Twitter, preprocessing Arabic text and Annotating the data set. After collecting and annotating the dataset, some data processing tasks are applied, such as feature selections, machine learning algorithms and deep neural networks. The efficiency of these methods are assessed and compared.

This paper continues in Section II with a brief survey of work on sentiment analysis in English and the Arabic languages. Section III details the process of collection, pre-processing and filtering that went into creating our data set; and then the annotating procedure. The use of deep neural networks and other machine learning methods, including text feature selection, on the dataset is described in Section IV. Finally, conclusions and ideas for future work are discussed in Section V.

## II. RELATED WORK

There are many studies on sentiment analysis and a variety of approaches have been developed. English has the greatest number of sentiment analysis studies, while research is more limited for other languages including Arabic. This section discusses several papers in the field of sentiment analysis using either English or Arabic.

Speriosu et al. [5] compared three different approaches, by using lexicon-based, maximum entropy classification and label propagation respectively. Several English datasets were used as training, evaluating and testing sets, and only positive and negative tweets were included in the datasets, whereas neutral tweets were eliminated. The experiment illustrated that the maximum entropy algorithm had a better result than the lexicon-based predictor and the accuracy improved for the test set of the polarity dataset from 58.1% to 62.9%. The label propagation obtained a better accuracy of 71.2%, by combining tweets and lexical features.

Kumar and Sebastian [6] presented a novel way to do the sentiment analysis on a Twitter data set in English, by extracting opinion words from the corpus. The types of words that were focused on, were adjectives, verbs and adverbs. Two methods were used: the corpus-based method for finding semantic of adjectives and the dictionary-based method for finding semantic of verbs and adverbs. After extracting opinion words, each word gets a score, whether it is positive, negative or neutral. The comprehensive tweet's score is measured by using individual score of each word using a linear equation.

Saif et al. [7] applied semantic features to the analysis of the Twitter users' opinions. There are three different English data sets used: Stanford Twitter Sentiment (STS), Health Care Reform (HCR) and Obama-McCain Debate (OMD). Three approaches were used to add semantic features for sentiment classification purposes, which were replacement, augmentation and interpolation. Baselines features like unigrams, parts of speech (POS), sentiment-topic and semantic sentiment analysis were used in the experiments. The best result in the experiments came from using the interpolation approach, unigram features and naïve Bayes classifier. The paper showed that large datasets are best analyzed by semantic methods, while sentiment-topic is the best method for small datasets or limited topics.

Shoukry and Rafea [8] addressed the Arabic sentence-level to perform sentiment analysis on 1000 tweets. Support Vector Machines (SVM) and Naïve Bayes (NB) were used in the experiment together with unigram and bigram text feature extraction. There were no differences between the results using different text feature extractions, but there were variations on the accuracy results using different classifiers: SVM ≈ 72% and NB ≈ 65%.

Ben Salamah and Elkhlifi [9] collected about 340,000 Arabic tweets about debates in the Kuwait National Assembly. The data was classified into positive and negative classes using decision trees (J48, alternating decision tree and random tree) and SVMs. The average precision results of the methods was 76% and the average recall was 61%.

Abdulla et al [10] created an Arabic dataset for sentiment analysis which contains 2000 tweets divided into positive, the first half, and negative, the second half. Two methods were applied to the dataset, which were corpus-based "Supervised Learning" and lexicon-based "Unsupervised Learning". Four supervised machine learning algorithms were applied, i.e., SVM, NB, D-Tree and K-Nearest Neighbor. The SVM and NB obtained better results, around 80%. On the other hand, the lexicon-based approach indicates that with a large lexicon the accuracy results were improving. There were three different phases, phase I has 1000 words, phase II has 2500 words and phase III has 3500 words. The accuracy started from 16.5% in phase one then it reached at 48.8% in phase two and it achieved 58.6% in phase three.

Kim [11] utilized convolutional neural networks to classify sentences using seven different English data sets, with the Movie Reviews dataset (introduced in [12]) being one of them. The experiment model built on top of "word2vec" [13] and various model variations were used. The experiments achieved good results of classifying sentences from different data sets.

## III. AN ARABC DATA SET ON HEALTH

The work described in this paper involved several steps, which are retrieving the data, filtering, pre-processing and annotating the data set, and finally applying some machine learning on the collected dataset. Collecting the data from Twitter using the Twitter API and defining keyword based queries related to health is the first step. The second step is a challenging one because the retrieved data has much noise and needs to be cleaned. Annotating the tweets in the data set to either positive of negative classes will occur after filtering it. After annotating the data set, several machine learning algorithms can be applied to the data set using different text features extractions. Figure I shows the workflow of this project.

FIGURE I. VISUALIZING THE WORK FLOW

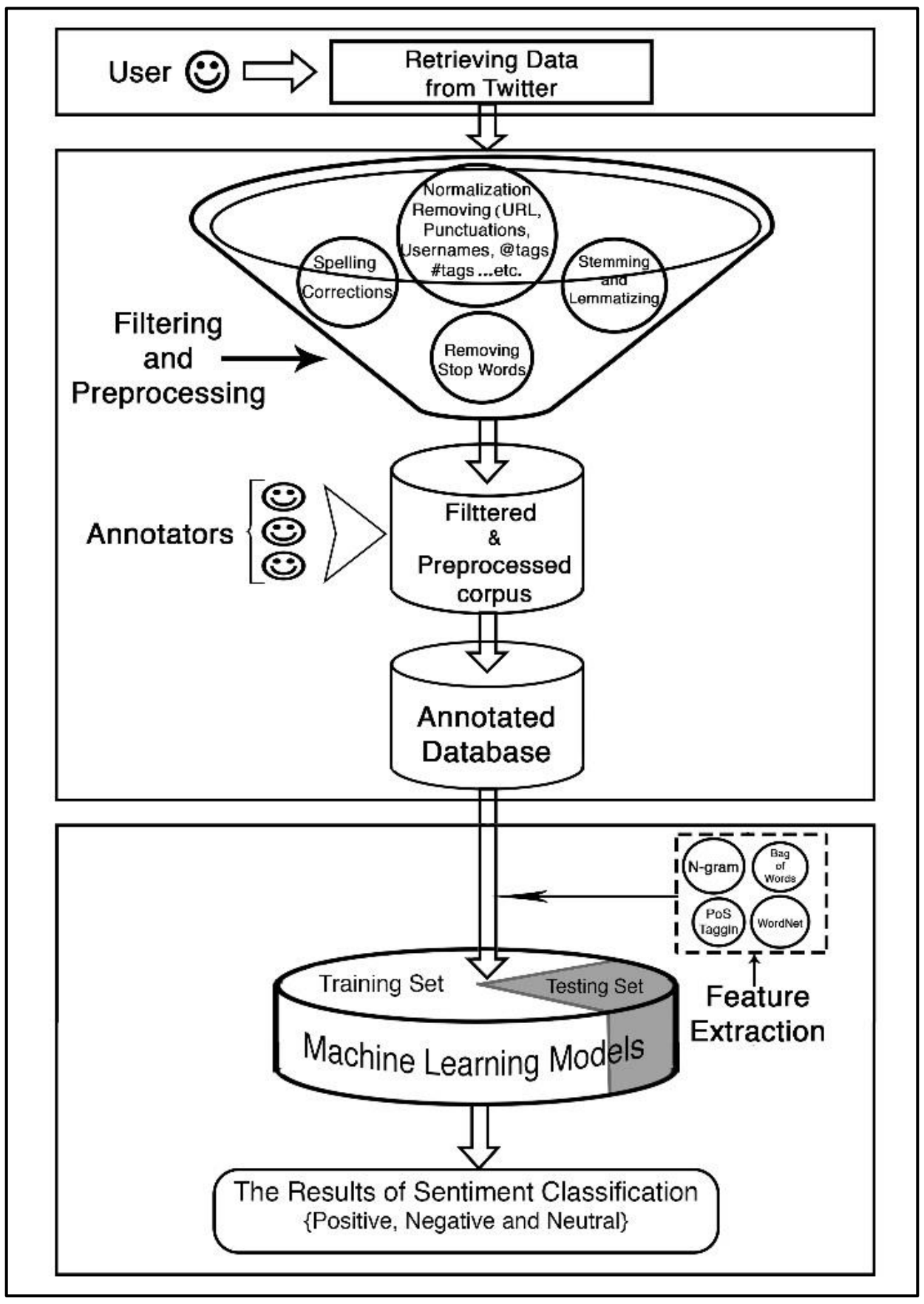

### A. *Data Collection*

The data was collected from 01/02/2016 to 31/07/2016 via Twitter. The first approach was to retrieve tweets using some general words related to health, such as "مستشفى, Hospital" , "مستوصف , Clinic" , "صحه , Health", etc. However, the majority of tweets found this was were not useful because they do not express any opinions, which is the aim of the study. Alternatively, observing trending hashtags, which are the most popular topics in Twitter at a specific time with many users involved in, was more useful. Three topics regarding to health were raised as trending hashtags and many users shared their opinions about them. They were as follows:

- #الصحة_تغلق_مستشفى

This topic is about closing a private hospital (Closing Hospital).

- #من_يعالج_الصحة

The meaning of this topic is asking a question about who will resolve the health problems (Solving Health).

- #ننتظر_تحسين_الصحة

This topic means that people are waiting for an improvement in the health services (Improving Health).

In addition to these topics, one topic was launched and asked users to post their opinions and experience about health services which was:

- #رأيك_بالخدمات_الصحية

This topic was launched especially for this study, which is about users' opinions regarding health services (Opinions about Health).

The number of retrieved tweets was massive (over 126 thousand) but it decreased to 2026 tweets after filtering and pre-processing. Table I shows the number of tweets of each topic before and after the pre-processing of the data.

TABLE I. THE CHANGES IN NUMBER OF TWEETS FOR EACH TOPIC BEFORE AND AFTER FILTERING THE DATASET

| Topics | Number of the tweets before the filtering | Number of the tweets after the filtering |
|---|---|---|
| **(Closing Hospital)** | 105275 tweets | 1009 tweets |
| **(Resolving Health)** | 11624 tweets | 492 tweets |
| **(Opinions about Health)** | 3033 tweets | 285 tweets |
| **(Improving Health)** | 7027 tweets | 240 tweets |
| **Total** | 126959 tweets | 2026 tweets |

### *B. Data Pre-processing and Normalization*

The number of collected tweets was sufficient to do the sentiment analysis experiment as it contains a variety of words and sentence structures. In contrast, the total number of tweets (126959 tweets) contained much noisy data and as the study focused on only positive or negative tweets, all noisy data was removed. The following points are examples of noisy data that were removed.

1) Spam tweets which are tweets that contain advertisements or harmful links [14].
2) Neutral tweets which do not have any opinions, such as news tweets.
3) Retweeted tweets, which start by "RT" [15].
4) Duplicated tweets, which were retrieved more than once.

In addition, as indicated in [15], some pre-processing steps were undertaken to the remaining tweets by removing:

1) Opinions unrelated to health.
2) Twitter users name which are like @user_name [15].
3) URLs which started by http:// until the next space, which indicates the end of the URL [15].
4) Some words like "available", "via" [15].
5) Hashtags topics.
6) Punctuations [15].

In addition to these tasks, normalization of some words or letters was performed as summarized below:

1) Removing Arabic short vowels (diacritics) " ِ, ً , ٌ, ٍ , َ , ُ , ْ , ّ" [16].
2) Removing the Tatweel character "ـ" which does not affect the meaning of the word [16].
3) Replacing the letter " ة " to the letter " ه " [16].
4) Replacing the letters " آ ، إ ، أ " to the letter " ا " [16].
5) Normalizing some words, especially words which contain the letter " Hamzah " " ئ ، ؤ " to one form because some users write it with " Hamzah " and other write it without it. For example the word " الطوارئ ، الطواري " can be written in two ways by users, but the words were normalized to one form which is "الطوارئ"
6) Normalizing any word with repeated letters, such as " كبيييييير " to be " كبير " [16].
7) Normalizing some special letters " ے , ک ", which are not Arabic letters, but they have the same shape of some Arabic letter.
8) Normalizing compound words using MS Excel by joining them by the character " _ "; such as some city names, for example "المدينه المنوره" which will be normalized to " المدينه_المنوره ".
9) Correcting words manually which were either missing some letters, replacing a letter by a wrong one, or writing the word in a wrong form.
10) Some Twitter users compress two words or more by ignoring the space between the words because of the characters limit on Twitter. These are normalized by manually returning the spaces between words.
11) Some users post their opinions in more than one tweet and the solution here was to combine them in one long tweet. After that, the length of combined tweet was reduced by removing unwanted words.

### *C. Data Anotating*

The data set has been annotated manually by three annotators and each tweet can be either positive or negative only. The reason of having three judges is to get three different opinions about each tweet, then calculating the majority vote of them ("The Mode"). There are eight rows in Table II for eight different situations of the annotators' classification. Also, the reasons of having only two classes are the difficulty of rating the opinions, the need of many annotators and the lack of scaled words in the corpus such as "very" in English language and "جدا" in the Arabic language.

Table II details the number of each different situation of the annotators' classification occurred in the dataset. For example, when all of them agree as positive or negative tweets, when two of them agree as positive and another one disagree and when two of them agree as negative and another one disagree. In addition to that, it shows the total number of positive and negative tweets.

TABLE II. SUMMARY OF ANOTATION PROCESS (POSITIVE = P, NEGATIVE = N)

| Annotator 1 | Annotator 2 | Annotator 3 | Total numbers of occurrences | Final Sentiment | Total |
|---|---|---|---|---|---|
| P | P | P | 502 times | P | 628 Positive Tweets |
| P | P | N | 49 times | P | |
| P | N | P | 74 times | P | |
| N | P | P | 3 times | P | |
| P | N | N | 135 times | N | 1398 Negative Tweets |
| N | P | N | 18 times | N | |
| N | N | P | 15 times | N | |
| N | N | N | 1230 times | N | |
| Total | | | 2026 tweets | | 2026 tweets |

From Table II the accuracies of each annotator can be measured. The accuracy of Annotator 1 is 93%, the accuracy of Annotator 2 is 95% and the accuracy of Annotator 3 is 97%.

Figure II shows the distributions of positive and negative tweets numbers per each annotator. It is clear that Annotators 2 and 3 have almost similar number of positive and negative tweets, but the Annotator 1 is slightly different. Overall, the data set is unbalanced, with the negative tweets more prevalent than the positive tweets.

FIGURE II. VISUALIZING THE NUMBER OF POSITIVE AND NEGATIVE TWEETS IN THE DATASET BASED ON THE THREE DIFFERENT ANNOTATORS

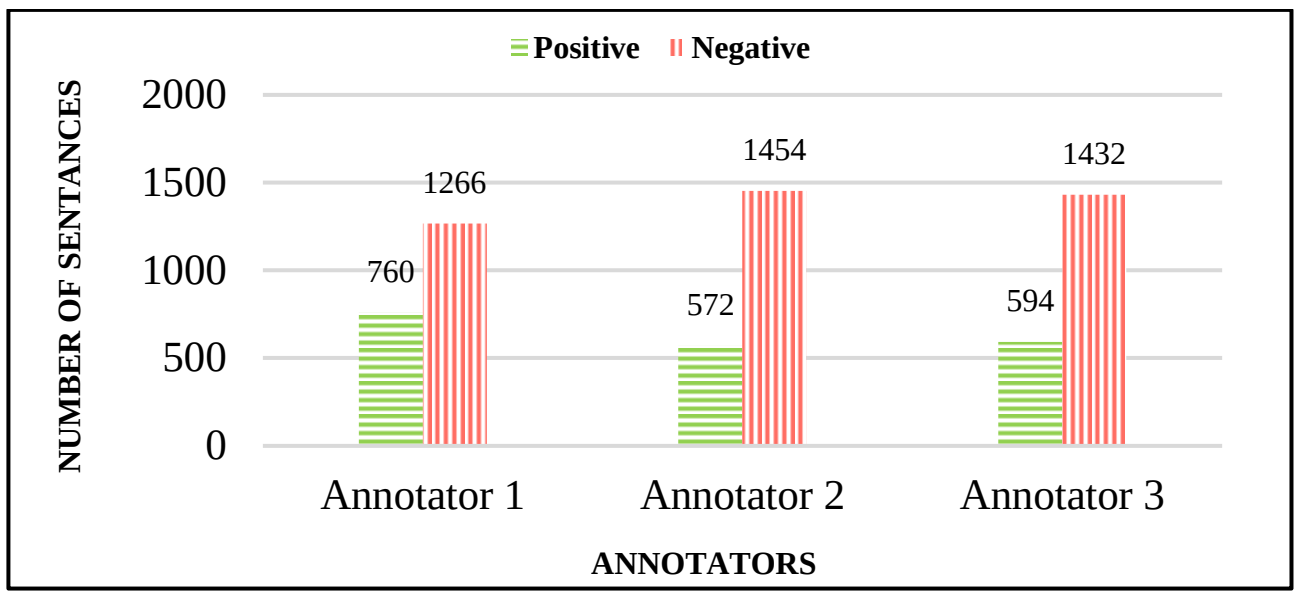

## IV. EXPERMENTS AND RESULTS

The objective of this experiment is to investigate the efficiency of utilizing Deep Neural Networks and other Machine Learning algorithms on a newly developed Arabic Health Services Data Set. [17] The efficiency of them can be measured by calculating the *Accuracy* of the classification task, which is defined as:

$$Accuracy = \frac{(TP+TN)}{(TP+TN+FP+FN)},$$

where (*TP*) is the number of true positives, (*TN*) is the number of true negatives, (*FP*) is the number of false positives and (*FN*) is the number of false negatives.

### A. *Machine Learning Algorithms*

In this section of the experiment, a combination of "Unigram" and "Bigram" techniques were used for text feature selection in this experiment. TF-IDF (Term Frequency and Inverse Document Frequency) [17] weighting words was used to weight each feature in the corpus and the maximum 1000 weighted features were fed to the machine learning algorithm. There are three machine learning algorithms that were used: Naïve Bayes (NB), Logistic Regression (LR) and Support Vector Machines (SVMs). [18] The NB algorithm used involved Multinomial Naive Bayes and Bernoulli Naive Bayes, and the SVMs used involved Support Vector Classification, Linear Support Vector Classification, Stochastic Gradient Descent and Nu-Support Vector Classification. The experiment had three phases by using different sizes for the training set and the testing set. In Phase I the training set was 60% of the data set and the testing set was 40% of the data set. In Phase II the testing set reduced to 30% and the training set was 70%. In Phase III the training set was increased 10% and the testing set was reduced 10%. Table III shows the results of all different classifiers in different phases using the previously explained text feature selection.

TABLE III. THE RESULTS OF 3 CLASSIFIERS THAT WERE USED WITH TF-IDF FEATURE SELECTION, "UNIGRAM" AND "BIGRAM"

| No. | Name of the Algorithm | Accuracy | | |
|---|---|---|---|---|
| | | Phase I | Phase II | Phase III |
| 1 | Multinomial Naive Bayes | 87.42 | 88.98% | 90.14% |
| 2 | Bernoulli Naive Bayes | 87.29% | 87.50% | 89.16% |
| 3 | Logistic Regression | 86.92% | 88.32% | 86.94% |
| 4 | Support Vector | 89.27% | 90.13% | 90.88% |
| 5 | Linear Support Vector | 89.39% | 90.46% | 91.37% |
| 6 | Stochastic Gradient Descent | 88.28% | 88.98% | 91.87% |
| 7 | Nu-Support Vector | 86.31% | 87.82% | 86.20% |

### B. *Deep Neural Networks Algorithms*

Deep learning is a popular approach in computational modeling today [19]. The model consists of a big number of hidden layers and neurons to represent the data with different abstractions. It works efficiently and effectively with large datasets. Neural Networks with many hidden layers, Convolutional Neural Networks and some Recurrent Neural Networks are examples of this [20].

In this section of the experiment, Deep and Convolutional Neural Networks were used. The Deep Neural Network model had three hidden layers and each layer has 1500 neurons. The input features used were 741 words, which were based on their frequency between 5 and 100 times in the corpus where the output of the model is either positive or negative. The data set was divided into 80% for training and 20% for testing. Figure III shows the obtained accuracy results of the experiment using Deep Neural Network, which reached about 85% in 500 epochs. Table IV shows the confusion matrix of the Deep Neural Network experiment on the test set. Also, it details the numbers of actual and predicted classes.

FIGURE III. ACCURACY RESULTS PER 500 EPOCHS FOR THE DEEP NEURAL NETWORK

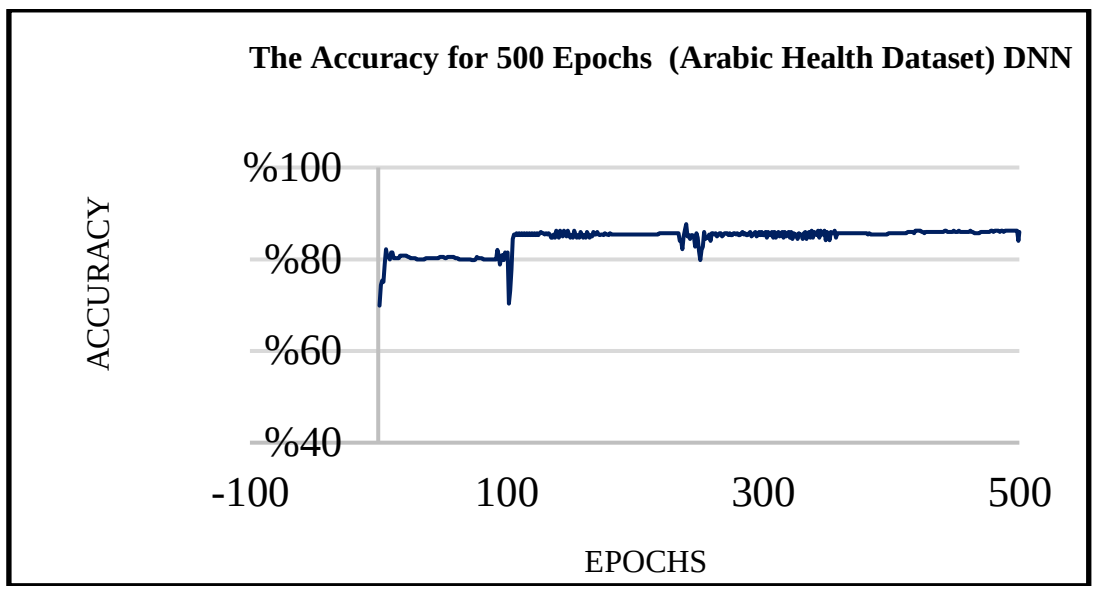

TABLE IV. DEEP NEURAL NETWORK CONFUSION MATRIX

| | | Predicted Classes | | Total |
|---|---|---|---|---|
| | | Negative | Positive | |
| Actual Classes | Negative | **265** | **14** | 279 |
| | Positive | **43** | **83** | 126 |
| Total | | 308 | 97 | 405 |

Convolutional Neural Networks (CNNs) were also used in the experiment. All the vocabulary in the corpus was trained using "word2vec" [13] to create the input vectors of the models. The sequence length of each vector was 52 because the longest sentence in the data set has 52 words. A (3, 4) sliding window were used to filter the size of the matrix and the number of epochs was 100. The data set was divided into 80% for training and 20% for testing. The accuracy obtained in this experiment was about 90%. Figure IV shows the accuracy results on 500 epochs by using a Convolutional Neural Network. Moreover, the confusion matrix was measured on the test set and it can be found in Table IV.

FIGURE IV. ACCURACY RESULTS PER 100 EPOCHS FOR THE CONVOLUTIONAL NEURAL NETWORK

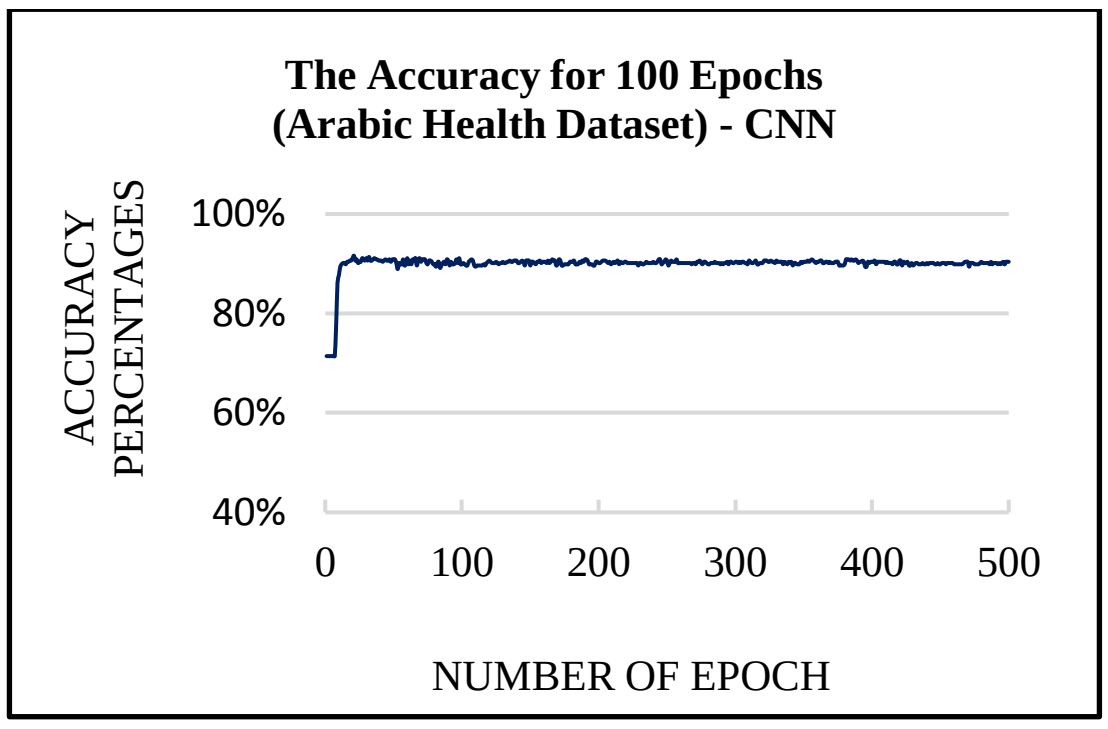

TABLE IV. CONVOLUTIONAL NEURAL NETWORK CONFUSION MATRIX

| | | Predicted Classes | | Total |
|---|---|---|---|---|
| | | Negative | Positive | |
| Actual Classes | Negative | **274** | **16** | 290 |
| | Positive | **23** | **93** | 116 |
| Total | | 297 | 109 | 406 |

## V. CONCLUSION AND FUTURE WORK

This paper introduces a new Arabic data set for sentiment analysis about health services. The paper also detailed the process of collecting Twitter tweets, the way of filtering, pre-processing Arabic text by removing unwanted data, removing some unrelated words and text and normalizing the text. Moreover, it explained the procedure of annotating the data set manually by three annotators. The initial experiments were conducted by utilizing Deep Neural Networks and several other Machine Learning algorithms. NB, LR, SVM, DNNs and CNNs were used and the accuracy in each experiment was recorded. The accuracy results were roughly between 85% and 91% and the best classifiers were SVM using Linear Support Vector Classification and Stochastic Gradient Descent. The SVM classifier accuracy is similar to the first annotator's accuracy.

There will be further studies and experiments on using different text features extraction and other Deep Neural Network and Recurrent Neural Network architectures in order to increase the accuracy of the results. In addition to this, the negation words in Arabic will be studied to increase the prediction performance and, as the data set is unbalanced, dealing with unbalanced data set techniques will be another topic for our future studies.